\PassOptionsToPackage{unicode}{hyperref}
\PassOptionsToPackage{hyphens}{url}
\PassOptionsToPackage{dvipsnames,svgnames,x11names}{xcolor}
\documentclass[
]{article}

\usepackage{amsmath,amssymb}
\usepackage{iftex}
\ifPDFTeX
  \usepackage[T1]{fontenc}
  \usepackage[utf8]{inputenc}
  \usepackage{textcomp} 
\else 
  \usepackage{unicode-math}
  \defaultfontfeatures{Scale=MatchLowercase}
  \defaultfontfeatures[\rmfamily]{Ligatures=TeX,Scale=1}
\fi
\usepackage{lmodern}
\ifPDFTeX\else  
\fi
\IfFileExists{upquote.sty}{\usepackage{upquote}}{}
\IfFileExists{microtype.sty}{
  \usepackage[]{microtype}
  \UseMicrotypeSet[protrusion]{basicmath} 
}{}
\makeatletter
\@ifundefined{KOMAClassName}{
  \IfFileExists{parskip.sty}{%
    \usepackage{parskip}
  }{
    \setlength{\parindent}{0pt}
    \setlength{\parskip}{6pt plus 2pt minus 1pt}}
}{
  \KOMAoptions{parskip=half}}
\makeatother
\usepackage{xcolor}
\ifx\paragraph\undefined\else
  \let\oldparagraph\paragraph
  \renewcommand{\paragraph}[1]{\oldparagraph{#1}\mbox{}}
\fi
\ifx\subparagraph\undefined\else
  \let\oldsubparagraph\subparagraph
  \renewcommand{\subparagraph}[1]{\oldsubparagraph{#1}\mbox{}}
\fi

\usepackage{color}
\usepackage{fancyvrb}

\DefineVerbatimEnvironment{Highlighting}{Verbatim}{commandchars=\\\{\}}
\usepackage{framed}
\definecolor{shadecolor}{RGB}{241,243,245}
\newenvironment{Shaded}{\begin{snugshade}}{\end{snugshade}}

\newcommand{\AttributeTok}[1]{\textcolor[rgb]{0.40,0.45,0.13}{#1}}

\newcommand{\CommentTok}[1]{\textcolor[rgb]{0.37,0.37,0.37}{#1}}

\newcommand{\DecValTok}[1]{\textcolor[rgb]{0.68,0.00,0.00}{#1}}

\newcommand{\FunctionTok}[1]{\textcolor[rgb]{0.28,0.35,0.67}{#1}}

\newcommand{\NormalTok}[1]{\textcolor[rgb]{0.00,0.23,0.31}{#1}}

\newcommand{\OtherTok}[1]{\textcolor[rgb]{0.00,0.23,0.31}{#1}}

\newcommand{\SpecialCharTok}[1]{\textcolor[rgb]{0.37,0.37,0.37}{#1}}

\newcommand{\StringTok}[1]{\textcolor[rgb]{0.13,0.47,0.30}{#1}}

\usepackage{longtable,booktabs,array}
\usepackage{calc} 
\usepackage{etoolbox}
\makeatletter
\patchcmd\longtable{\par}{\if@noskipsec\mbox{}\fi\par}{}{}
\makeatother
\IfFileExists{footnotehyper.sty}{\usepackage{footnotehyper}}{\usepackage{footnote}}
\makesavenoteenv{longtable}
\usepackage{graphicx}
\makeatletter
\def\maxwidth{\ifdim\Gin@nat@width>\linewidth\linewidth\else\Gin@nat@width\fi}
\def\maxheight{\ifdim\Gin@nat@height>\textheight\textheight\else\Gin@nat@height\fi}
\makeatother
\setkeys{Gin}{width=\maxwidth,height=\maxheight,keepaspectratio}
\makeatletter
\def\fps@figure{htbp}
\makeatother
\newlength{\cslhangindent}
\newlength{\csllabelwidth}
\newlength{\cslentryspacingunit} 
\newenvironment{CSLReferences}[2] 
 {
  \setlength{\parindent}{0pt}
  \ifodd #1
  \let\oldpar\par
  \def\par{\hangindent=\cslhangindent\oldpar}
  \fi
  \setlength{\parskip}{#2\cslentryspacingunit}
 }%
 {}
\usepackage{calc}

\usepackage{arxiv}
\usepackage{orcidlink}
\usepackage{amsmath}
\usepackage[T1]{fontenc}
\makeatletter
\@ifpackageloaded{caption}{}{\usepackage{caption}}
\AtBeginDocument{%
\ifdefined\contentsname
  \renewcommand*\contentsname{Table of contents}
\else
  \newcommand\contentsname{Table of contents}
\fi
\ifdefined\listfigurename
  \renewcommand*\listfigurename{List of Figures}
\else
  \newcommand\listfigurename{List of Figures}
\fi
\ifdefined\listtablename
  \renewcommand*\listtablename{List of Tables}
\else
  \newcommand\listtablename{List of Tables}
\fi
\ifdefined\figurename
  \renewcommand*\figurename{Figure}
\else
  \newcommand\figurename{Figure}
\fi
\ifdefined\tablename
  \renewcommand*\tablename{Table}
\else
  \newcommand\tablename{Table}
\fi
}
\@ifpackageloaded{float}{}{\usepackage{float}}
\floatstyle{ruled}
\@ifundefined{c@chapter}{\newfloat{codelisting}{h}{lop}}{\newfloat{codelisting}{h}{lop}[chapter]}
\floatname{codelisting}{Listing}

\makeatother
\makeatletter
\@ifpackageloaded{caption}{}{\usepackage{caption}}
\@ifpackageloaded{subcaption}{}{\usepackage{subcaption}}
\makeatother
\makeatletter
\@ifpackageloaded{tcolorbox}{}{\usepackage[skins,breakable]{tcolorbox}}
\makeatother
\makeatletter
\@ifundefined{shadecolor}{\definecolor{shadecolor}{rgb}{.97, .97, .97}}
\makeatother
\ifLuaTeX
  \usepackage{selnolig}  
\fi
\IfFileExists{bookmark.sty}{\usepackage{bookmark}}{\usepackage{hyperref}}
\IfFileExists{xurl.sty}{\usepackage{xurl}}{} 
\hypersetup{
  pdftitle={rollama: An R package for using generative large language models through Ollama},
  pdfkeywords={R, large language models, open models},
  colorlinks=true,
  linkcolor={blue},
  filecolor={Maroon},
  citecolor={Blue},
  urlcolor={Blue},
  pdfcreator={LaTeX via pandoc}}

\newcommand{\runninghead}{A Preprint }
\title{rollama: An R package for using generative large language models
through Ollama}
\def\asep{\\\\\\ } 
\author{\textbf{Johannes B.
Gruber}~\orcidlink{0000-0001-9177-1772}\\\\VU
Amsterdam\\\\\asep\textbf{Maximilian
Weber}~\orcidlink{0000-0002-1174-449X}\\\\Johannes Gutenberg University
Mainz\\\\}
\date{}
\begin{document}
\maketitle
\begin{abstract}
\texttt{rollama} is an R package that wraps the \texttt{Ollama} API,
which allows you to run different Generative Large Language Models
(GLLM) locally. The package and learning material focus on making it
easy to use \texttt{Ollama} for annotating textual or imagine data with
open-source models as well as use these models for document embedding.
But users can use or extend \texttt{rollama} to do essentially anything
else that is possible through OpenAI's API, yet more private,
reproducible and for free.
\end{abstract}
{\bfseries \emph Keywords}
\def\sep{\textbullet\ }
R \sep large language models \sep 
open models

\ifdefined\Shaded\renewenvironment{Shaded}{\begin{tcolorbox}[borderline west={3pt}{0pt}{shadecolor}, boxrule=0pt, frame hidden, breakable, enhanced, interior hidden, sharp corners]}{\end{tcolorbox}}\fi

\hypertarget{statement-of-need}{%
\section{Statement of need}\label{statement-of-need}}

As researchers embrace the next revolution in computational social
science, the arrival of GLLM\footnote{Also referred to generative AI or
  Generative Pre-trained Transformer (GPT).}, there is a critical need
for open-source alternatives (Spirling 2023). This need arises to avoid
falling into a reproducibility trap or becoming overly dependent on
services offered by for-profit companies.

After the release of ChatGPT, researchers started using OpenAI's API for
annotating text with GPT models (e.g., Gilardi, Alizadeh, and Kubli
2023; He et al. 2023). However, this approach presents several
shortcomings, including privacy and replication issues associated with
relying on proprietary models over which researchers have no control
whatsoever (Spirling 2023; Weber and Reichardt 2023).

Fortunately, since GLLMs were popularized by OpenAI's ChatGPT a little
more than a year ago, a large and active alliance of open-source
communities and technology companies has made considerable efforts to
provide open models that rival, and sometimes surpass, proprietary ones
(Alizadeh et al. 2023; Irugalbandara et al. 2024).

One method of utilizing open models involves downloading them from a
platform known as Hugging Face and using them via Python scripts.
However, there is now software available that facilitates access to
these models in an environment, allowing users to simply specify the
model or models they wish to use. This can be done locally on one's
computer, and the software is called \texttt{Ollama}. The R package
\texttt{rollama} provides a wrapper to use the \texttt{Ollama} API
within R.

\hypertarget{background-ollama}{%
\section{\texorpdfstring{Background:
\texttt{Ollama}}{Background: Ollama}}\label{background-ollama}}

\texttt{Ollama} can be installed using dedicated installers for macOS
and Windows, or through a bash installation script for Linux\footnote{\url{https://ollama.com/download}}.
However, our preferred method is to utilize Docker -- an open-source
containerization tool. This approach enhances security and simplifies
the processes of updating, rolling back, and completely removing
\texttt{Ollama}. For convenience, we provide a Docker compose file to
start a container running \texttt{Ollama} and Open WebUI -- a browser
interface strongly inspired by ChatGPT -- in a GitHub Gist\footnote{\url{https://gist.github.com/JBGruber/73f9f49f833c6171b8607b976abc0ddc}}.

\hypertarget{usage}{%
\section{Usage}\label{usage}}

After \texttt{Ollama} is installed, the R-package \texttt{rollama} can
be installed from CRAN (the Comprehensive R Archive Network):

\begin{Shaded}
\begin{Highlighting}[]
\FunctionTok{install.packages}\NormalTok{(}\StringTok{"rollama"}\NormalTok{)}
\end{Highlighting}
\end{Shaded}

or from GitHub using remotes:

\begin{Shaded}
\begin{Highlighting}[]
\CommentTok{\# install.packages("remotes")}
\NormalTok{remotes}\SpecialCharTok{::}\FunctionTok{install\_github}\NormalTok{(}\StringTok{"JBGruber/rollama"}\NormalTok{)}
\end{Highlighting}
\end{Shaded}

After that, the user should check whether the \texttt{Ollama} API is up
and running.

\begin{Shaded}
\begin{Highlighting}[]
\FunctionTok{library}\NormalTok{(rollama)}
\FunctionTok{ping\_ollama}\NormalTok{()}
\end{Highlighting}
\end{Shaded}

After installation, the first step is to pull one of the models from
\href{https://ollama.com/library}{ollama.com/library} by using the model
tag. By calling \texttt{pull\_model()} without any arguments will
download the default model.

\begin{Shaded}
\begin{Highlighting}[]
\CommentTok{\# pull the default model}
\FunctionTok{pull\_model}\NormalTok{()}
\CommentTok{\# pull a specified model by providing the model tag}
\FunctionTok{pull\_model}\NormalTok{(}\StringTok{"gemma:2b{-}instruct{-}q4\_0"}\NormalTok{)}
\end{Highlighting}
\end{Shaded}

\hypertarget{main-functions}{%
\subsection{Main functions}\label{main-functions}}

The core of the package are two functions: \texttt{query()} and
\texttt{chat()}. The difference is that \texttt{chat()} saves the
history of the conversation with an \texttt{Ollama} model, while
\texttt{query()} treats every question as a new conversation.

To ask a single question, the \texttt{query()} function can be used.

\begin{Shaded}
\begin{Highlighting}[]
\FunctionTok{query}\NormalTok{(}\StringTok{"why is the sky blue?"}\NormalTok{)}
\end{Highlighting}
\end{Shaded}

For conversational interactions, the \texttt{chat()} function is used.

\begin{Shaded}
\begin{Highlighting}[]
\FunctionTok{chat}\NormalTok{(}\StringTok{"why is the sky blue?"}\NormalTok{)}
\FunctionTok{chat}\NormalTok{(}\StringTok{"and how do you know that?"}\NormalTok{)}
\end{Highlighting}
\end{Shaded}

\hypertarget{reproducible-outcome}{%
\subsection{Reproducible outcome}\label{reproducible-outcome}}

Within the \texttt{model\_params} of the \texttt{query()} function,
setting a seed is possible. When a seed is used, the temperature must be
set to ``0'' to ensure consistent output for repeated prompts.

\begin{Shaded}
\begin{Highlighting}[]
\FunctionTok{query}\NormalTok{(}\StringTok{"Why is the sky blue? Answer in one sentence."}\NormalTok{,}
      \AttributeTok{model\_params =} \FunctionTok{list}\NormalTok{(}\AttributeTok{seed =} \DecValTok{42}\NormalTok{, }\AttributeTok{temperature =} \DecValTok{0}\NormalTok{))}
\end{Highlighting}
\end{Shaded}

\hypertarget{examples}{%
\section{Examples}\label{examples}}

We present several examples to illustrate some functionalities. As
mentioned previously, many tasks that can be performed through OpenAI's
API can also be accomplished by using open models within
\texttt{Ollama}. Moreover, these models can be controlled with a seed,
ensuring reproducible results.

\hypertarget{annotating-text}{%
\subsection{Annotating text}\label{annotating-text}}

For annotating text data, various prompting strategies are available
(see Weber and Reichardt 2023 for a comprehensive overview). These
strategies primarily differ in whether or how many examples are given
(Zero-shot, One-shot, or Few-shot) and whether reasoning is involved
(Chain-of-Thought).

When writing a prompt we can give the model content for the system part,
user part and assistant part. The system message typically includes
instructions or context that guides the interaction, setting the stage
for how the user and the assistant should interact. For an annotation
task we could write: ``You assign texts into categories. Answer with
just the correct category.'' The following example is a zero-shot
approach only containing a system message and one user message. In
practice, annotating a single text is rarely the goal. For guidance on
batch annotations\footnote{\url{https://jbgruber.github.io/rollama/articles/annotation.html\#batch-annotation}}
and working with dataframes\footnote{\url{https://jbgruber.github.io/rollama/articles/annotation.html\#another-example-using-a-dataframe}},
refer to the package documentation.

\begin{Shaded}
\begin{Highlighting}[]
\FunctionTok{library}\NormalTok{(tibble)}
\FunctionTok{library}\NormalTok{(purrr)}
\NormalTok{q }\OtherTok{\textless{}{-}} \FunctionTok{tribble}\NormalTok{(}
  \SpecialCharTok{\textasciitilde{}}\NormalTok{role,    }\SpecialCharTok{\textasciitilde{}}\NormalTok{content,}
  \StringTok{"system"}\NormalTok{, }\StringTok{"You assign texts into categories. Answer with just the correct category."}\NormalTok{,}
  \StringTok{"user"}\NormalTok{,   }\StringTok{"text: the pizza tastes terrible}\SpecialCharTok{\textbackslash{}n}\StringTok{categories: positive, neutral, negative"}
\NormalTok{)}
\FunctionTok{query}\NormalTok{(q, }\AttributeTok{model\_params =} \FunctionTok{list}\NormalTok{(}\AttributeTok{seed =} \DecValTok{42}\NormalTok{, }\AttributeTok{temperature =} \DecValTok{0}\NormalTok{))}
\end{Highlighting}
\end{Shaded}

\hypertarget{using-multimodal-models}{%
\subsection{Using multimodal models}\label{using-multimodal-models}}

\texttt{Ollama} also supports multimodal models, which can interact with
(but at the moment not create) images. A model designed for image
handling, such as the llava model, needs to be pulled. Using
\texttt{pull\_model("llava")} will download the model.

\begin{Shaded}
\begin{Highlighting}[]
\FunctionTok{pull\_model}\NormalTok{(}\StringTok{"llava"}\NormalTok{)}
\FunctionTok{query}\NormalTok{(}\StringTok{"Excitedly desscribe this logo"}\NormalTok{, }\AttributeTok{model =} \StringTok{"llava"}\NormalTok{,}
      \AttributeTok{images =} \StringTok{"https://ollama.com/public/ollama.png"}\NormalTok{)}
\end{Highlighting}
\end{Shaded}

\hypertarget{obtaining-text-embeddings}{%
\subsection{Obtaining text embeddings}\label{obtaining-text-embeddings}}

\texttt{Ollama}, and hence rollama, can be utilized to generate text
embeddings. In short, text embedding uses the knowledge of the meaning
of words inferred from the context they are used in and saved in a large
language model through its training to turn text into meaningful vectors
of numbers (Kroon et al. 2023). This technique is a powerful
preprocessing step for supervised machine learning and often increases
the performance of a classification model substantially (Laurer et al.
2024). GLLMs can produce high-quality document embeddings, such as the
current default model of \texttt{Ollama} llama2. However, to speed up
the procedure, one can use specialized embedding models like
nomic-embed-text\footnote{\url{https://ollama.com/library/nomic-embed-text}}
or all-minilm\footnote{\url{https://ollama.com/library/all-minilm}}.

\begin{Shaded}
\begin{Highlighting}[]
\FunctionTok{pull\_model}\NormalTok{(}\AttributeTok{model =} \StringTok{"nomic{-}embed{-}text"}\NormalTok{)}
\FunctionTok{embed\_text}\NormalTok{(}\AttributeTok{text =} \StringTok{"It’s a beautiful day"}\NormalTok{, }\AttributeTok{model =} \StringTok{"nomic{-}embed{-}text"}\NormalTok{)}
\end{Highlighting}
\end{Shaded}

Transformer based text embedding models were previously only available
in Python, with the only pathway for R users to employ them being the
Python interface provided by \texttt{reticulate} (Kalinowski et al.
2024) or packages built on top of it (e.g. Chan 2023; Kjell, Giorgi, and
Schwartz 2023). While these usually work well, non-technical R users
regularly struggle with the requirement to set up the necessary Python
environments. \texttt{Ollama}, and hence \texttt{rollama}, circumvents
this step by packaging the whole application in an executable or Docker
image. For a more detailed example of using embeddings for training
supervised machine learning models and doing classification tasks, refer
to the package documentation\footnote{\url{https://jbgruber.github.io/rollama/articles/text-embedding.html}}.

\hypertarget{learning-material}{%
\section{Learning material}\label{learning-material}}

We provide tutorials for the package at
\href{https://jbgruber.github.io/rollama/}{jbgruber.github.io/rollama}
and an initial overview is available as a YouTube video\footnote{\url{https://youtu.be/N-k3RZqiSZY}}.

\hypertarget{references}{%
\section*{References}\label{references}}
\addcontentsline{toc}{section}{References}

\hypertarget{refs}{}
\begin{CSLReferences}{1}{0}
\leavevmode\vadjust pre{\hypertarget{ref-alizadeh2023opensource}{}}%
Alizadeh, Meysam, Maël Kubli, Zeynab Samei, Shirin Dehghani, Juan Diego
Bermeo, Maria Korobeynikova, and Fabrizio Gilardi. 2023. {``Open-Source
Large Language Models Outperform Crowd Workers and Approach ChatGPT in
Text-Annotation Tasks.''} arXiv.
\url{https://doi.org/10.48550/arXiv.2307.02179}.

\leavevmode\vadjust pre{\hypertarget{ref-chan_grafzahl2023}{}}%
Chan, Chung-Hong. 2023. {``Grafzahl: Fine-Tuning Transformers for Text
Data from Within r.''} \emph{Computational Communication Research} 5
(1): 76--84. \url{https://doi.org/10.5117/CCR2023.1.003.CHAN}.

\leavevmode\vadjust pre{\hypertarget{ref-GilardiChatGPT2023}{}}%
Gilardi, Fabrizio, Meysam Alizadeh, and Maël Kubli. 2023. {``ChatGPT
Outperforms Crowd Workers for Text-Annotation Tasks.''}
\emph{Proceedings of the National Academy of Sciences} 120 (30).
\url{https://doi.org/10.1073/pnas.2305016120}.

\leavevmode\vadjust pre{\hypertarget{ref-He_Lin_et_al._2023}{}}%
He, Xingwei, Zhenghao Lin, Yeyun Gong, A.-Long Jin, Hang Zhang, Chen
Lin, Jian Jiao, Siu Ming Yiu, Nan Duan, and Weizhu Chen. 2023.
{``AnnoLLM: Making Large Language Models to Be Better Crowdsourced
Annotators.''} arXiv. \url{https://doi.org/10.48550/arXiv.2303.16854}.

\leavevmode\vadjust pre{\hypertarget{ref-irugalbandara_trade-off_2024}{}}%
Irugalbandara, Chandra, Ashish Mahendra, Roland Daynauth, Tharuka
Kasthuri Arachchige, Krisztian Flautner, Lingjia Tang, Yiping Kang, and
Jason Mars. 2024. {``A Trade-Off Analysis of Replacing Proprietary LLMs
with Open Source SLMs in Production.''} arXiv.
\url{https://doi.org/10.48550/arXiv.2312.14972}.

\leavevmode\vadjust pre{\hypertarget{ref-ushey_reticulate2024}{}}%
Kalinowski, Tomasz, Kevin Ushey, J. J. Allaire, Yuan Tang, Dirk
Eddelbuettel, Bryan Lewis, Sigrid Keydana, Ryan Hafen, and Marcus
Geelnard. 2024. \emph{Reticulate: Interface to 'Python'}.
\url{https://CRAN.R-project.org/package=reticulate}.

\leavevmode\vadjust pre{\hypertarget{ref-kjell_text2023}{}}%
Kjell, Oscar, Salvatore Giorgi, and Andrew H. Schwartz. 2023. {``The
Text-Package: An r-Package for Analyzing and Visualizing Human Language
Using Natural Language Processing and Deep Learning.''}
\emph{Psychological Methods}. \url{https://doi.org/10.1037/met0000542}.

\leavevmode\vadjust pre{\hypertarget{ref-kroon_transfer-learning2023}{}}%
Kroon, Anne, Kasper Welbers, Damian Trilling, and Wouter van Atteveldt.
2023. {``Advancing Automated Content Analysis for a New Era of Media
Effects Research: The Key Role of Transfer Learning.''}
\emph{Communication Methods and Measures} 0 (0): 1--21.
\url{https://doi.org/10.1080/19312458.2023.2261372}.

\leavevmode\vadjust pre{\hypertarget{ref-laurer_benchmarking2024}{}}%
Laurer, Moritz, Wouter van Atteveldt, Andreu Casas, and Kasper Welbers.
2024. {``Less Annotating, More Classifying: Addressing the Data Scarcity
Issue of Supervised Machine Learning with Deep Transfer Learning and
BERT-NLI.''} \emph{Political Analysis} 32 (1): 84--100.
\url{https://doi.org/10.1017/pan.2023.20}.

\leavevmode\vadjust pre{\hypertarget{ref-Spirling_2023}{}}%
Spirling, Arthur. 2023. {``Why Open-Source Generative AI Models Are an
Ethical Way Forward for Science.''} \emph{Nature} 616 (7957): 413--13.
\url{https://doi.org/10.1038/d41586-023-01295-4}.

\leavevmode\vadjust pre{\hypertarget{ref-weber2023evaluation}{}}%
Weber, Maximilian, and Merle Reichardt. 2023. {``Evaluation Is All You
Need. Prompting Generative Large Language Models for Annotation Tasks in
the Social Sciences. A Primer Using Open Models.''} arXiv.
\url{https://doi.org/10.48550/arXiv.2401.00284}.

\end{CSLReferences}

\end{document}